\documentclass{article}

\usepackage[final]{neurips_2023}

\usepackage[utf8]{inputenc}
\usepackage[T1]{fontenc}
\usepackage{hyperref}
\usepackage{url}
\usepackage{booktabs}
\usepackage{amsfonts}
\usepackage{nicefrac}
\usepackage{microtype}
\usepackage{graphicx}
\usepackage{xcolor}
\usepackage{lipsum}
\usepackage{caption}
\usepackage{subcaption}
\usepackage{amsmath}
\usepackage{booktabs}

\title{
 Can LLMs \textit{understand} Math?\\Exploring the Pitfalls in Mathematical Reasoning
}

\author{
  \makebox[\textwidth][c]{%
    Tiasa Singha Roy$^*$, \quad
    Aditeya Baral$^*$, \quad
    Ayush Rajesh Jhaveri, \quad
    Yusuf Baig
  } \\
  New York University \\
  \texttt{\{ts5478, ab12057, aj4332, yb2510\}@nyu.edu}
}

\begin{document}

\maketitle

\begin{abstract}

Large language models (LLMs) demonstrate considerable potential in various natural language tasks but face significant challenges in mathematical reasoning, particularly in executing precise, multi-step logic. However, current evaluation frameworks judge their performance solely based on accuracy, which only accounts for the final answer. This study explores these pitfalls by employing a novel evaluation framework. We propose an evaluation metric called the MAPLE score, which holistically quantifies reasoning misalignment by integrating error rates, redundancy, and validity.

\end{abstract}

\section{Introduction}
Large Language Models (LLMs) have demonstrated impressive capabilities across tasks such as text generation, language translation, question answering, and sentiment analysis. However, their performance diminishes in complex reasoning tasks, particularly within the mathematical domain. While LLMs perform adequately on elementary math problems, they often need help with tasks requiring precise, step-by-step reasoning, leading to errors in solution validity and logical consistency. These limitations underscore the need for a holistic evaluation of their mathematical reasoning abilities to identify weaknesses and guide targeted improvements.

This study introduces a multi-stage evaluation methodology to systematically assess LLMs' mathematical reasoning capabilities. We prompt various LLMs to generate solutions using the MATH dataset. The process involves iterative self-reflection to evaluate reasoning steps, identify misalignments, and compile error labels such as calculation errors, misinterpretations, and incoherent outputs.

We propose the MAPLE (Mathematical Pitfalls and Logical Evaluation) score as a holistic metric to quantify reasoning misalignment. This score incorporates error rates, redundancy, and validity, offering a comprehensive evaluation. Our findings reveal patterns of errors and limitations across different mathematical topics and levels, providing insights into the challenges LLMs face in complex reasoning tasks.
\section{Related Work}


We draw on insights from ReasonEval\cite{xia2024evaluating}, which argues that solely relying on final answer accuracy can mask the use of unnecessary or incorrect intermediate steps in the mathematical reasoning process. It introduces a methodology highlighting the importance of going beyond accuracy in evaluating LLM performance for mathematical reasoning. To extend this methodology, we leverage ideas of self-reflection\cite{shinn2024reflexion}, which proposes a method for using LLMs to self-correct themselves for reasoning. We take motivation from this method to use the LLM to identify pitfalls and patterns in its reasoning evaluation. However, these methods depend on external sources for effective self-improvement. Our work builds on this by using oracle labels directly within the LLM, allowing it to identify and analyze patterns in its reasoning failures autonomously. Furthermore, while past studies, such as \cite{huang2023large}\cite{kamoi2024can}, argue that using Oracle labels for self-correction may not be realistic for all applications, we propose employing them here in a self-feedback context.

\section{Approach}

\begin{figure}[h]
     \centering
     \includegraphics[width=0.7\textwidth]{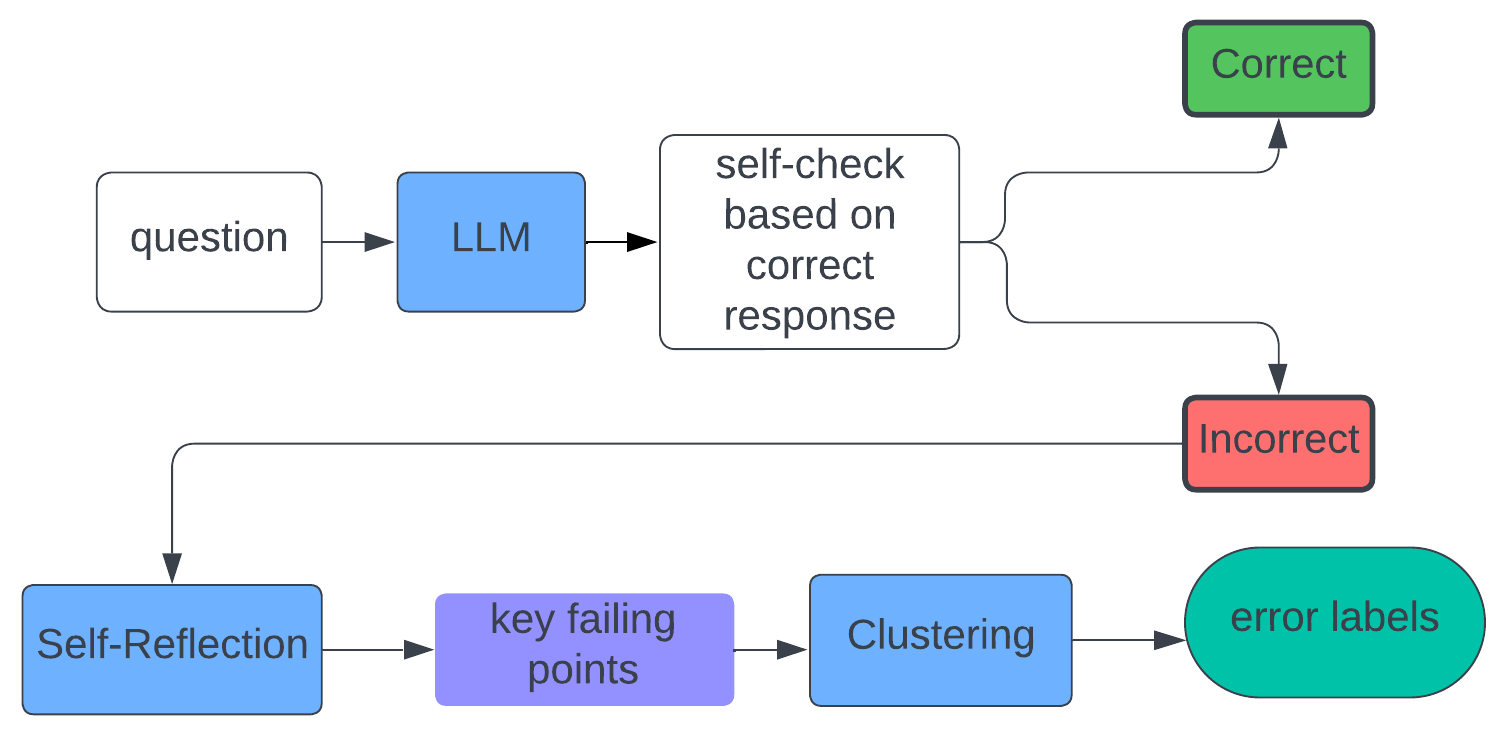}
     \caption{\textbf{Architecture of our LLM Agent evaluation and identification of errors}. The LLM's generated answer is evaluated in a multi-turn set-up to identify the failing points in the generated response using \textit{self-reflection} and \textit{clustering}.}
     \label{fig:approach-flow}
\end{figure}

\subsection{Stage 1 - Evaluating the Final Answer and Approach}

As shown in Figure \ref{fig:approach-flow}, we initialize the LLM agent to generate an answer, $a'_i$ based on the initial prompt $p_i$ and question $q_i$. For each $a'_i$, we create a multi-turn setup, which includes providing the agent with the correct solution to induce self-checking. We provide the LLM with the correct final answer value $a_{fi}$ to check. This approach verifies whether the final answers $(a'_{fi}, a_{fi})$ match to identify incorrect cases for the LLM agent.

For these incorrect responses, we invoke self-reflection for the incorrect samples. This step uses generated and correct response pair $(a'_i, a_i)$ to return a generation analysis which highlights the points of misalignment of the reasoning steps with the actual solution. We use BERT\cite{devlin2018bert} based embeddings to wrap these failing points and perform clustering to compile a consistent set of error labels $L$ to encompass issues across the samples. 


\begin{figure}[h]
     \centering
     \includegraphics[width=0.5\textwidth]{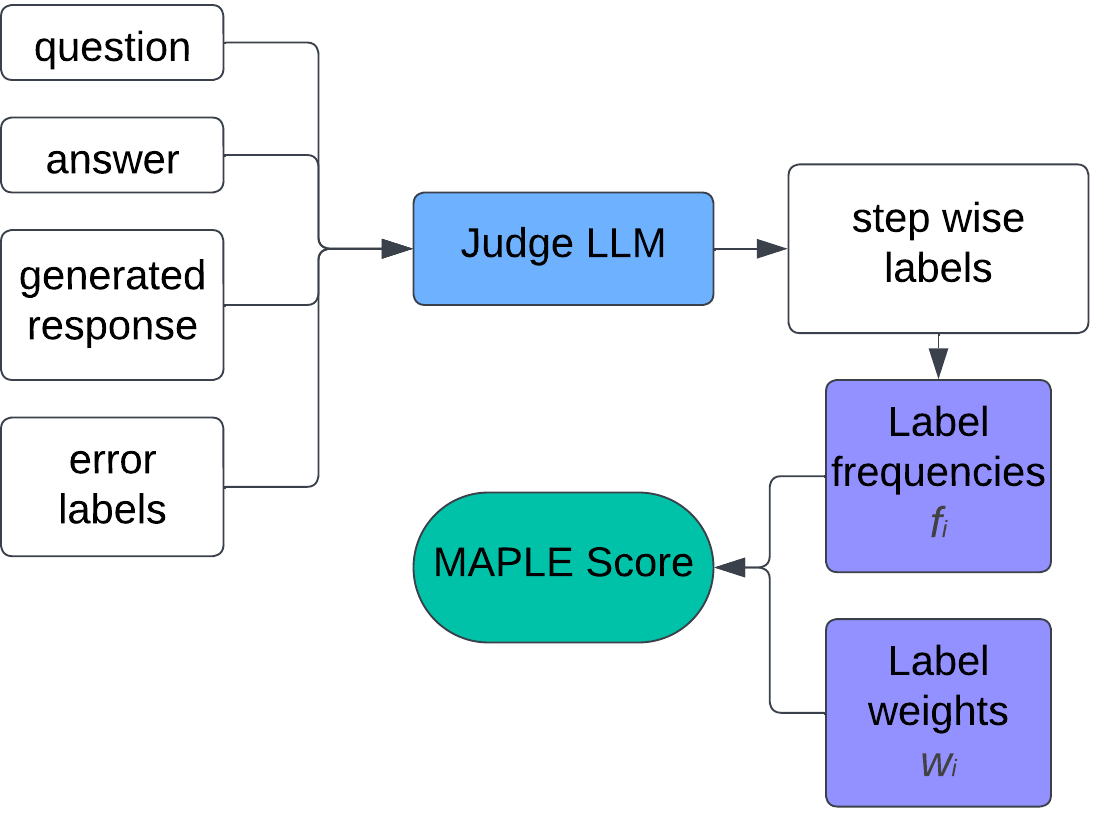}
     \caption{\textbf{Architecture of our Judge LLM Agent and MAPLE Score}. The Judge LLM provides step-wise analysis to compute MAPLE score using \textit{label-frequencies} and \textit{label-weights}.}
     \label{fig:approach-flow2}
\end{figure}

\subsection{Stage 2 - LLM as a Judge and Computing Incorrectness}



While the compiled error labels $L$ help provide insights into the broad classification of mathematical mistakes made while solving a problem, it is also crucial to identify and correlate each mathematical reasoning step with these labels.

As shown in Figure \ref{fig:approach-flow2}, we initialize a judge LLM and prompt it with the error labels $L$, the question $q_i$, the correct solution $a_i$, and the generated solution $a'_i$ to create a set of error labels $S$ corresponding to each erroneous reasoning step. These step-wise error labels are then used to compute the \textit{degree of incorrectness} in the incorrect solution.


\subsection{Stage 3 - Computing MAPLE Score}


Given a step-wise collection of error labels ${S} = [{[{l_1, l_2, ..., l_n}],[{l_1, l_2, ..., l_n}], ...}]$, we first compute the frequency $f_l$ of each label $l \in L$ . This represents the relevance of a particular label to an incorrectly generated sample. We consider the logarithm of frequencies instead of raw frequencies to reduce sensitivity to large frequency values, with a $+1$ added term to ensure numerical stability when no errors are made.

We further compute the error rate $e$ as the weighted average of the frequencies $f_l$ weighted by the penalty score for each label $w_l$. The penalty score for each label, as shown in section \ref{sec:error-label-penalty-weights}, was aggregated over the results of a human survey which ranked the error labels in increasing order of incorrectness. The penalty score allows us to individually weigh the contribution of each error label to compute the MAPLE score.

\begin{equation}
    \label{eq:error-rate}
    e = \frac{\sum_{l \in L} w_l \cdot log(1 + f_l)}{\sum_{l \in L} w_l}
\end{equation}

The redundancy $r$ and validity $v$ of the overall solution are computed using ReasonEval\cite{xia2024evaluating} within the range of values $r, v \in [0, 1]$. The MAPLE score decreases with an increase in validity and increases with an increase in redundancy of the solution. Finally, we express the error metric $e$ by applying the $tanh$ function such that $\text{MAPLE}_{score} \in [0, 1]$.

\begin{equation}
    \label{eq:MAPLE}
    \text{MAPLE}_{score} = tanh \left(\frac{e \cdot v}{r}\right)
\end{equation}

\section{Experiments}

\subsection{Data}
We use the MATH \cite{hendrycksmath2021} dataset, which comprises 12,500 competition mathematics problems. The problems vary in complexity from levels 1 through 5 and span mathematics categories, consisting of Intermediate Algebra, Precalculus, Algebra, Prealgebra, Geometry, Counting \& Probability, and Number Theory.

\subsection{Experiment Setup}
We choose the four models for our study from the four popular LLM families, namely Gemini, GPT-4, Llama, and Mixtral. More details on the models used can be found in section \ref{sec:experiment-setup}.

\subsection{Evaluation method}

To evaluate the reliability of the MAPLE score, we perform a thorough validation process on the judge LLM predictions. We manually annotate a representative sample from the MATH dataset, consisting of incorrect mathematical answers generated by various LLM families. A multi-label approach is employed to comprehensively capture the types of errors present in the responses. The error labels predicted by the judge LLM are then compared against these human-annotated labels. This alignment accuracy can be found in section \ref{sec:evaluation-llm-judge}
.


\subsection{Results}

\subsubsection{Error Classification}
Based on the clustering of the LLM self-reflections for incorrect answers, we obtained the following error labels:

\begin{enumerate}
    \item \textbf{Complete misunderstanding}. The model completely fails to understand the question and its requirements.
    \item \textbf{Partial misunderstanding}. The model partially fails to understand the question or its requirements.
    \item \textbf{Incorrect Method}. The model applies a concept with the correct formula but it is unrelated to the given question.
    \item \textbf{Incorrectly Applied Method}. The model chooses the right concept with the correct formula which is related to the given question and can be used to solve it, but applied it incorrectly.
    \item \textbf{Calculation Error}. Errors in arithmetic calculations.
    \item \textbf{Incoherent Output}. Junk text with repeated characters or phrases.
    \item \textbf{No Solution}. Failure to reach a final answer.
\end{enumerate}

The error labels are provided as prompts to the judge LLM, which identifies the errors present in the mathematical answers generated. These identified errors are subsequently used for computing the MAPLE score.

\subsubsection{MAPLE Score Computation}
\begin{figure}[ht]
    \centering
    \includegraphics[width=1\textwidth]{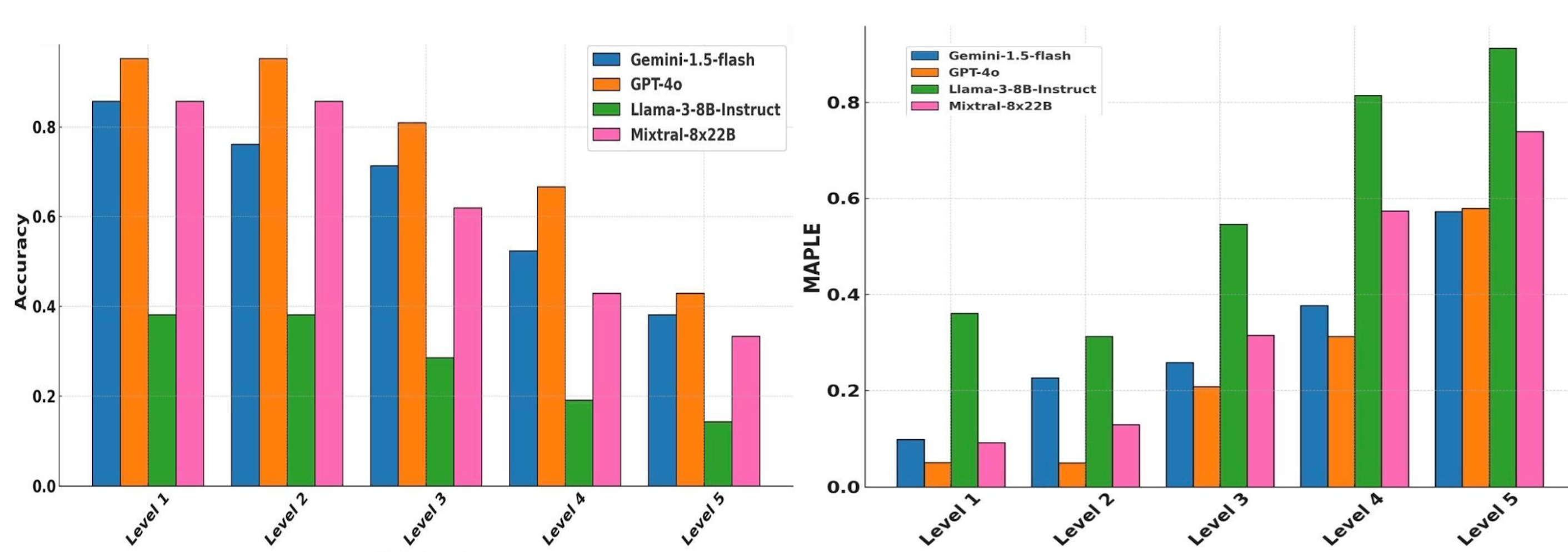}
    \caption{\textbf{Comparison of LLM performance across difficulty levels on the MATH Dataset}. Level 1 represents the easiest and Level 5 represents the toughest math problems. We observe a correlation between final answer accuracy and the degree of incorrectness represented by the MAPLE score.}
    \label{fig:math-error-class}
\end{figure}

We evaluated the mathematical answers generated by various LLMs for the MATH dataset using our proposed approach. The results, categorized by difficulty level, are presented in Figure \ref{fig:math-error-class}. The left graph demonstrates that as the difficulty level increases, accuracy declines across all models. Conversely, the right graph shows that the MAPLE score rises with increased difficulty, with the highest MAPLE score observed for the Llama model. This suggests that the Llama model exhibits the most significant issues in mathematical reasoning.

Additionally, we performed a topic-wise analysis of the LLM-generated answers, the results of which are provided in section \ref{sec:topic-wise-results}.

\section{Future Work}
Future efforts will expand the evaluation framework to include a broader range of error types, such as topic-specific reasoning issues, and incorporate ranking of error labels for more nuanced scoring. Addressing hallucination in LLMs through fine-tuning for evaluation-specific tasks and exploring alternatives to LLMs as judges will enhance alignment with human judgment. Testing the framework on diverse models, datasets, and interdisciplinary reasoning tasks will validate its robustness. Additionally, refining methods to reduce redundancy and improve logical coherence in reasoning steps will be critical for advancing LLMs' mathematical problem-solving capabilities.

\bibliographystyle{unsrt}
\bibliography{references}

\newpage
\appendix
\section{Appendix}

\subsection{Error Label Penalty Weights}
\label{sec:error-label-penalty-weights}
\begin{table}[h!]
\centering
\caption{Error Label Penalty Weights}
\label{sec:error-label-penalty-weights-table}
\begin{tabular}{@{}lc@{}}
\toprule
\textbf{Error Label}                 & \textbf{Penalty Weight} \\ \midrule
Complete Misunderstanding           & 0.95                    \\
Partial Misunderstanding            & 0.75                    \\
Incorrectly Applied Method          & 0.40                    \\
Calculation Error                   & 0.10                    \\
Incoherent Output                   & 1.00                    \\
No Solution                         & 1.00                    \\ \bottomrule
\end{tabular}
\end{table}

\subsection{Experiment Setup}
\label{sec:experiment-setup}

The four LLMs chosen for our study are Gemini-1.5-Flash from Gemini, GPT-4o from GPT-4, Llama-3-8B-Instruct\footnote{\href{https://huggingface.co/mistralai/https://huggingface.co/meta-llama/Meta-Llama-3-8B-Instruct}{\texttt{https://huggingface.co/meta-llama/Meta-Llama-3-8B-Instruct}}} from LlaMa, and Mixtral-8x22B\footnote{\href{https://huggingface.co/mistralai/Mixtral-8x22B-Instruct-v0.1}{\texttt{https://huggingface.co/mistralai/Mixtral-8x22B-Instruct-v0.1}}} from Mistral. All fine-tuned open-source models were taken from HuggingFace, while respective APIs were used for GPT-4o and Gemini-1.5-Flash. Outputs were generated with the temperature parameters set to \texttt{1} for Gemini-1.5-Flash and GPT-4o, \texttt{0.3} for Mixtral-8x22B, and \texttt{0.05} for Llama-3-8B-Instruct.

To cluster the failing points and compile a set of error labels $L$, we use MathBERT\cite{peng2021mathbertpretrainedmodelmathematical}\footnote{\href{https://huggingface.co/tbs17/MathBERT}{\texttt{https://huggingface.co/tbs17/MathBERT}}}, a BERT model fine-tuned on math word problems. Similarly, we use Mathstral-7B\footnote{\href{https://huggingface.co/mistralai/Mathstral-7B-v0.1}{\texttt{https://huggingface.co/mistralai/Mathstral-7B-v0.1}}}, a fine-tuned Mistral model for mathematical problem solving from HuggingFace for our Judge LLM. 

\subsection{Evaluation of LLM as Judge}
\label{sec:evaluation-llm-judge}
\begin{figure}[ht]
    \centering
    \includegraphics[width=1\textwidth]{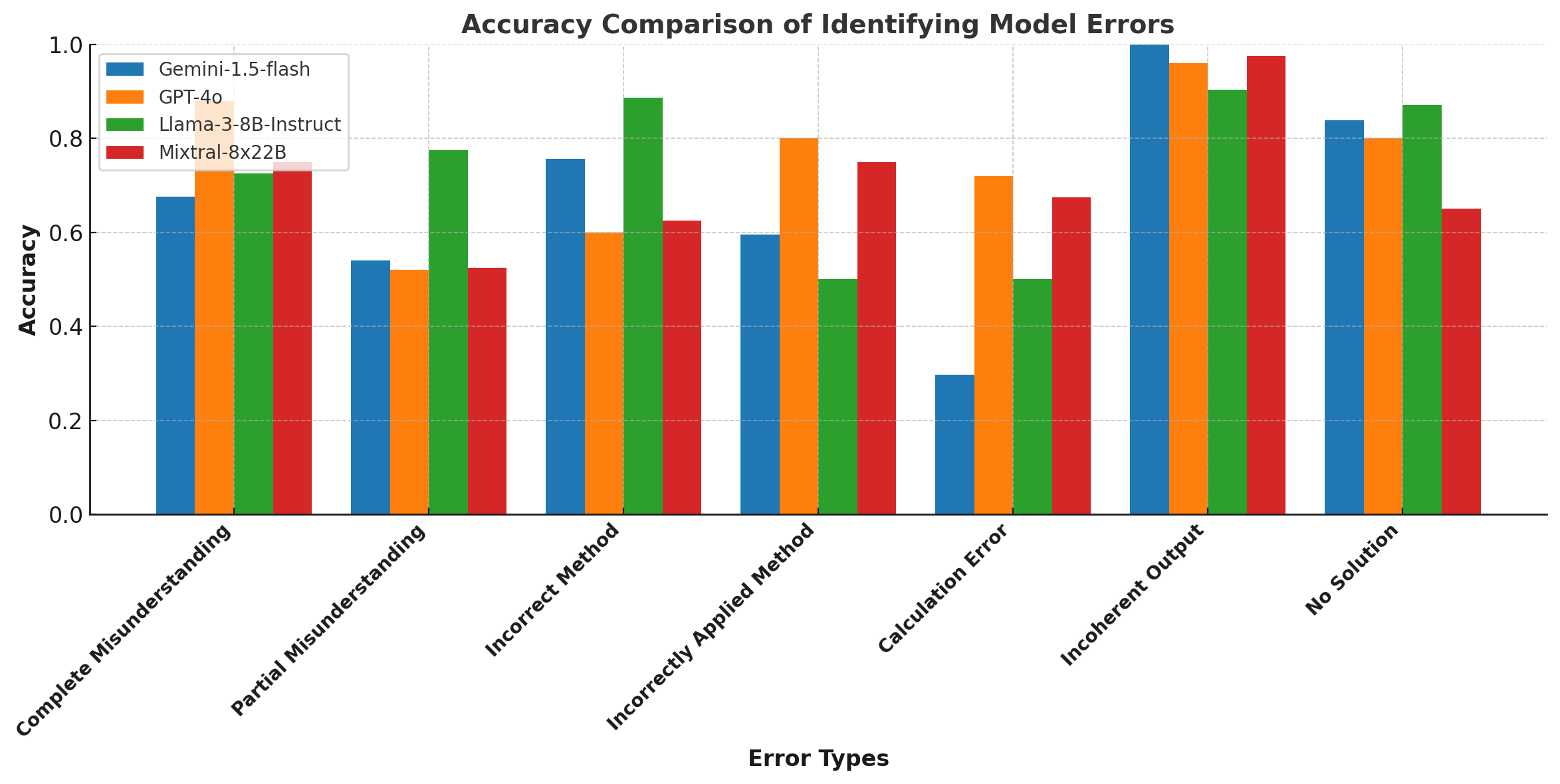}
    \caption{\textbf{Comparison of accuracy of the LLM as a Judge in predicting error labels for generated solutions}. We observe that most predictions match human annotations for a representative sample of 105 evenly-distributed examples across difficulty levels and topics.}
    \label{fig:math-error-class2}
\end{figure}
\newpage

\subsection{Topic-wise Evaluation Scores}
\label{sec:topic-wise-results}
\begin{figure}[h]
     \centering
     \includegraphics[width=1\textwidth]{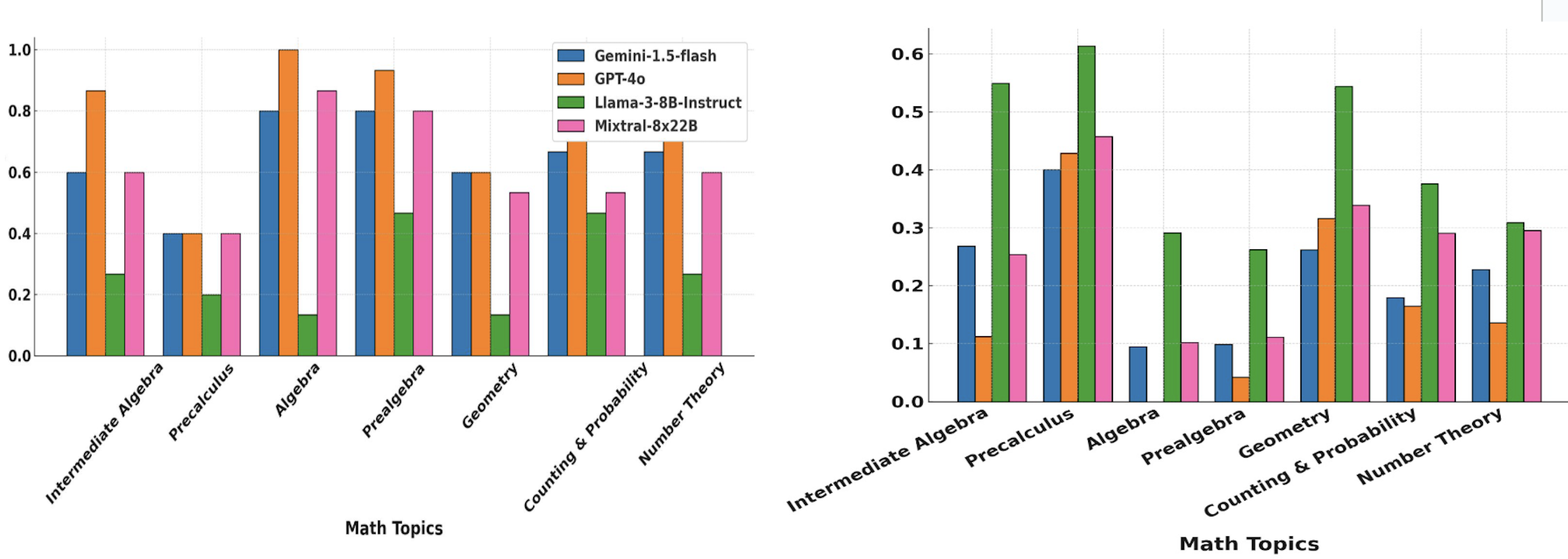}
     \caption{\textbf{Comparison of LLM performance across math topics on the MATH Dataset}. We observe that most models perform better at easier topics such as geometry while underperforming at tougher topics such as calculus.}
     \label{fig:approach-flow4}
\end{figure}
\end{document}